\documentclass[10pt,twocolumn,letterpaper]{article}

\usepackage{cvpr}              

\usepackage{times}
\usepackage{epsfig}
\usepackage{graphicx}
\usepackage{amsmath}
\usepackage{amssymb}
\usepackage{algorithm}
\usepackage{algorithmic}
\usepackage{booktabs}
\usepackage{multirow}
\usepackage{tabularx}
\usepackage{caption}
\usepackage{lineno}
\usepackage{xfp}
\usepackage{pifont}

\definecolor{cvprblue}{rgb}{0.21,0.49,0.74}
\usepackage[pagebackref,breaklinks,colorlinks,allcolors=cvprblue]{hyperref}

\def\paperID{15487} 
\def\confName{CVPR}
\def\confYear{2026}

\title{INTERLACE: Interleaved Layer Pruning and Efficient Adaptation in Large Vision-Language Models}

\author{Parsa Madinei \qquad Ryan Solgi \qquad Ziqi Wen \qquad Jonathan Skaza \\
Miguel Eckstein \qquad Ramtin Pedarsani\\
\textsc{UC Santa Barbara}\\
}

\begin{document}
\twocolumn[{%
\renewcommand\twocolumn[1][]{#1}%
\maketitle
\begin{center}
    \centering
    \includegraphics[width=1\textwidth]{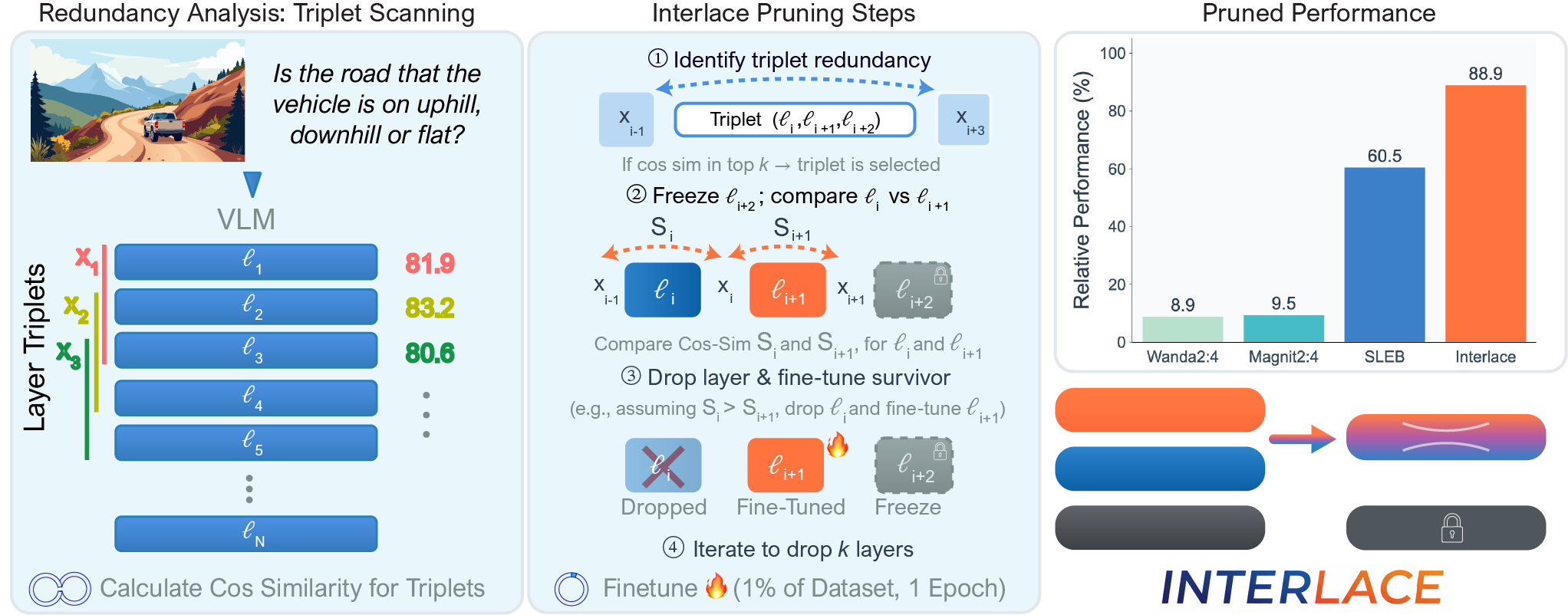}
    \captionof{figure}{\textbf{Interlace} is a layer pruning framework for VLMs. It first identifies local redundancy by calculating cosine similarity for ``triplets" of layers. In each selected triplet, the most redundant of the first two layers is dropped, the other is fine-tuned, and the third is frozen to act as a stable anchor. The performance comparison (top right) shows that \textit{Interlace} outperforms alternative pruning methods by 28.4\%.}
    \label{fig:teaser}
\end{center}%
\vspace{1em} 
}]
\begin{abstract}
\\[-18pt]
We introduce \textbf{Interlace}, a novel framework that prunes redundant layers in VLMs while maintaining performance through sample-efficient finetuning. Existing layer pruning methods lead to significant performance drop when applied to VLMs. Instead, we analyze triplets of consecutive layers to identify local redundancy, removing the most redundant of the first two layers, finetune the remaining layer to compensate for the lost capacity, and freeze the third layer to serve as a stable anchor during finetuning. We found that this interleaved finetune-freeze design enables rapid convergence with minimal data after pruning. By finetuning only a subset of layers on \textbf{just 1\%} of the FineVision dataset for \textbf{one epoch}, Interlace achieves \textbf{88.9\%} average performance retention after dropping 25\% of the network, achieving SOTA performance. Our code is available at: \url{https://github.com/pmadinei/Interlace.git} 
\end{abstract}
\section{Introduction}

The rapid advancement of Large Vision-Language Models (LVLMs) has transformed multimodal understanding, enabling unprecedented capabilities in visual question answering, image captioning, and multimodal reasoning. Models such as Qwen3-VL \cite{yang2025qwen3}, LLaVA-OneVision \cite{li2024llava}, and InternVL \cite{chen2024internvl} have achieved state-of-the-art performance by scaling both model capacity and training data. However, this scaling strategy introduces a critical inefficiency. To cater to diverse deployment needs, leading models are often released as a family of different-sized versions. The standard practice is to pretrain \textit{each} of these variants independently from scratch on the complete, massive dataset. This ``train-from-scratch'' paradigm for every model size dramatically multiplies the already immense computational demand, consuming vast amounts of memory and processing power. Fine-tuning methods have become a common practice to avoid full training costs; yet even fine-tuning \twocolumn[{%
\renewcommand\twocolumn[1][]{#1}%
\begin{center}
    \centering
    \includegraphics[width=0.9\textwidth]{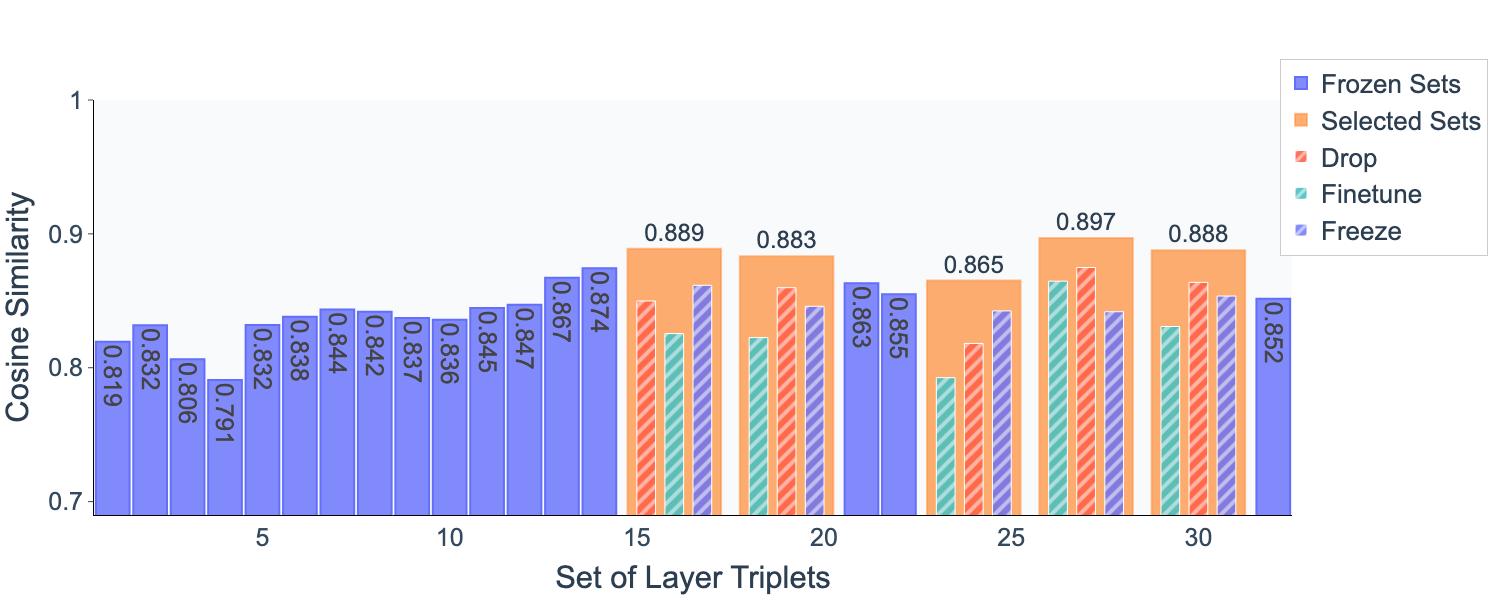}
    \captionof{figure}{\textbf{Triplet selection and layer assignment} based on cosine similarity scores in Qwen3-VL-8B. Selected triplets are highlighted with their layer assignments: the individual layer with the highest similarity score between the first two for dropping (red), the other layer in the first two for fine-tuning (cyan), and the last layer for freezing (blue). Unselected triplets remain frozen. Individual layer similarity scores within selected triplets are normalized to fit within the triplet's overall similarity range.
}
    \label{fig:similarities}
\end{center}%
}] \setlength{\parindent}{0pt} LVLMs is often challenging when available resources are limited. This computational burden poses critical challenges for deploying LVLMs in real-world applications, particularly in resource-constrained settings. Model compression techniques have emerged as a crucial solution to bridge this deployment gap~\cite{solgi2025saten}. Traditional approaches include quantization, knowledge distillation, and various forms of network pruning \cite{zhu2024survey}. Among these, layer pruning presents a particularly attractive avenue for LVLMs, as removing entire layers provides actual inference speedup without requiring specialized hardware support. Recent work on layer pruning for Large Language Models (LLMs) has demonstrated that certain layers contribute minimally to model output, suggesting inherent redundancy in deep transformer architectures \cite{elhoushi2024layerskip, fan2019reducing}. However, the application of these techniques to LVLMs leads to a significant performance drop \cite{sung2023ecoflap, solgi2025pgsvd}. Furthermore, some of these approaches require pretraining the model with randomly dropped layers, which prevents their application to standard, off-the-shelf pretrained models that were not trained using this specific methodology.

The key challenge in layer pruning for LVLMs lies in identifying which layers to remove and how to recover lost performance efficiently. Previous approaches, such as Streamlining \cite{chen2024streamlining}, prune consecutive layers based on individual layer importance scores in LLMs. However, this method creates large blocks of adjacent, modified layers, which leads to significant fine-tuning instability and poor convergence when applied to LVLMs \cite{ma2025short}, where inter-modal dependencies are more fragile and deeply integrated across layers. Moreover, fine-tuning strategies that update all remaining parameters prove both computationally expensive and sample-inefficient \cite{zhao2024pruning}, limiting practical applicability.

In this work, we present \textsc{Interlace}, a framework that addresses these convergence and efficiency challenges through three key innovations as illustrated in Fig.~\ref{fig:teaser}. First, we introduce a triplet-based layer importance metric that evaluates sets of three consecutive layers, capturing local redundancy patterns while avoiding the fine-tuning instability caused by removing large consecutive blocks. Second, we implement a strategic freeze-and-finetune approach where we freeze the last layer in each selected triplet. This frozen layer acts as a critical structural constraint, forcing the model to leverage original features and proving essential for stabilizing and accelerating convergence. Third, within each triplet's first two layers, we selectively prune the more redundant layer and fine-tune the remaining layer to compensate for the capacity of both, creating a final, interleaved architecture. This constrained, interleaved design enables efficient training with minimal data while maintaining robust performance. We fine-tuned a subset of layers after pruning on only 1\% of FineVision \cite{wiedmann2025finevision} for just 1 epoch and were able to achieve an average of 92.1\% of baseline accuracy on a wide range of challenging VLM generation tasks.

Our main contributions are:
\vspace{-5pt}
\begin{itemize}
\item We propose a novel triplet-based layer analysis framework that identifies non-consecutive redundant layers in LVLMs by evaluating cosine similarity across sets of three consecutive layers, avoiding the limitations of consecutive transformer block pruning strategies.

\item We introduce a selective freeze-and-finetune training recipe that constrains model exploration by freezing strategically selected baseline layers while fine-tuning compensatory layers, achieving superior performance with exceptional sample and memory efficiency. 

\item By fine-tuning a subset of layers on only \textit{1\%} of the FineVision dataset for just \textit{a single epoch}, we achieved state-of-the-art average performance retention with a score of \textbf{88.9\%} across 12 diverse VLM generation benchmarks after dropping 25\% of layers.

\item We provide comprehensive ablation studies comparing four alternative layer selection and training strategies, empirically validating each component of our framework.
\end{itemize}




\section{Related Work}

\paragraph{Layer Pruning in LLMs} Pruning has been extensively studied for LLM compression. Recent work has explored layer-level redundancy as a compression target. LLM-Streamline \cite{chen2024streamlining} introduced pruning consecutive layers based on cosine similarity of hidden states and trains a lightweight pruned network. The Shortened LLaMA \cite{kim2024shortened} demonstrated that depth pruning achieves better inference speedups than width pruning on resource-constrained devices. ShortGPT \cite{men2024shortgpt} showed that layers in LLMs exhibit more redundancy than previously expected. SLEB ~\cite{song2024sleb} pruned LLMs by removing entire transformer blocks. However, these approaches focus exclusively on language-only models and either prune consecutive layers or require extensive retraining. Sheared-LLaMA \cite{xia2024sheared} demonstrated that continued pretraining on pruned architectures can recover performance, but at substantial computational cost. Our work extends these insights to vision-language models while introducing a more efficient triplet-based selection strategy that avoids consecutive pruning and minimizes training requirements.

\textbf{Weight Pruning in LLMs.} Although some pruning methods drop entire components such as layers or transformer blocks~\citep{song2024sleb}; others result in partially prune weight matrices~\citep{back2025magnitude, ma2023llmpruner}. LLM-Pruner \cite{ma2023llmpruner} proposed structured pruning using gradient information. Wanda \cite{sun2023simple} prunes weights based on magnitude and activation norms without retraining, demonstrating that effective sparse subnetworks exist in LLMs. Magnitude attention-based dynamic pruning (MAP) ~\cite{back2025magnitude} leverages weight importance for both pruning and training. SparseGPT \cite{frantar2023sparsegpt} performs one-shot pruning using second-order information. However, these weight-level pruning methods face challenges in vision-language models: they require careful per-layer sparsity configuration, may not achieve actual inference speedups without specialized hardware, and often necessitate extensive retraining~\cite{sung2023ecoflap}. Our layer-level approach provides guaranteed speedups through architectural changes while requiring minimal fine-tuning on small datasets, offering a complementary and more deployment-friendly compression strategy for LVLMs.

\textbf{Pruning in LVLMs.} Studies on pruning in LVLMs has primarily focused on visual token reduction rather than model parameter pruning. FastV \cite{chen2024fastv} prunes visual tokens in early layers based on attention patterns, while PyramidDrop \cite{xing2024pyramiddrop} implements progressive token reduction across model depth. LLaVA-PruMerge \cite{shang2024llavaprumerge} combines token pruning and merging for adaptive compression. CoViPAL \cite{xing2024covipal} employs layer-wise contextualized pruning of visual tokens with plug-and-play modules. Recent work has also explored dynamic approaches: Dynamic-LLaVA \cite{tian2024dynamic} sparsifies both vision and language contexts across inference stages, while ATP-LLaVA \cite{liu2024atpllava} adaptively determines instance-specific token pruning ratios. These token-level methods complement but differ fundamentally from our layer-level approach, as they maintain the full model architecture while reducing input dimensionality. Notably, direct application of LLM layer pruning techniques to LVLMs has shown limited success. The work closest to ours explores layer importance in multimodal contexts \cite{kim2024compression}, demonstrating that vision-language tasks are more sensitive to layer removal than language-only tasks, motivating our careful triplet-based selection strategy.

\section{Methods}

\subsection{Preliminaries and Motivation}

Large Vision-Language Models typically consist of three primary components: a vision encoder (e.g., CLIP-ViT), a projection module for modality alignment, and an LLM decoder. Given an input image $\mathbf{I}$ and text prompt $\mathbf{T}$, the vision encoder produces visual features $\mathbf{V} = \text{Enc}(\mathbf{I}) \in \mathbb{R}^{n_v \times d}$, which are projected to the language model's embedding space via $\mathbf{V'} = \text{Proj}(\mathbf{V}) \in \mathbb{R}^{n_v \times d_{\text{llm}}}$. These visual embeddings are concatenated with text embeddings $\mathbf{E}_t = \text{Embed}(\mathbf{T})$ to form the input sequence $\mathbf{X}_0 = [\mathbf{V'}; \mathbf{E}_t]$ for the language model. The language model consists of $L$ transformer decoder layers, where each layer $\ell$ transforms the hidden states: $\mathbf{X}_\ell = f_\ell(\mathbf{X}_{\ell-1}; \theta_\ell)$. Each layer comprises multi-head self-attention and feed-forward networks with residual connections. The output logits are computed via $\mathbf{Y} = \text{LMHead}(\mathbf{X}_L)$ for next-token prediction.

Our key observation is that the post-pruning fine-tuning process becomes highly unstable when large blocks of consecutive layers are modified simultaneously, a common outcome in many pruning methods \cite{chen2024streamlining}. We found that attempting to fine-tune the resulting deep stack of adjacent, trainable layers leads to significant convergence challenges. The model fails to converge rapidly because these layers, lacking the stable, intermediate representations previously provided by their neighbors, struggle to adapt. Our key insight is to introduce a structural constraint: by strategically freezing layers and interleaving them between the layers selected for fine-tuning, we provide \textit{feature anchors}. This design forces the model to adapt within a more constrained feature space, which dramatically stabilizes the fine-tuning process and accelerates convergence even with minimal data.

\subsection{Triplet-Based Layer Importance}

We propose analyzing layers in triplets to capture local redundancy patterns while avoiding the limitations of consecutive pruning and fine-tuning. For a model with $L$ layers, we define $L-2$ overlapping triplets: $\mathcal{T}_i = \{\ell_i, \ell_{i+1}, \ell_{i+2}\}$ for $i \in [1, L-2]$. For each triplet, we compute the average cosine similarity across all tokens in the hidden states before and after the triplet:

\begin{equation}
S_{\text{triplet}}(i) = \frac{1}{N} \sum_{j=1}^{N} \cos(\mathbf{x}^{(j)}_{i-1}, \mathbf{x}^{(j)}_{i+2})
\end{equation}


\setlength{\parindent}{0pt}where $\mathbf{x}^{(j)}_{\ell}$ denotes the hidden state of token $j$ after layer $\ell$, and $N$ is the total number of tokens. We compute this metric using 10\% of the fine-tuning dataset (0.1\% of FineVision). Additionally, for each individual layer within these triplets, we compute its single-layer importance $S_{\text{layer}}(\ell)$, which is defined as the average token-wise cosine similarity between the hidden states before and after that specific layer. This dual-level analysis enables us to identify both locally redundant triplets and the least important layer within each triplet.

\subsection{Layer Selection Strategy}

Given a target pruning ratio $\rho$ (e.g., 0.10, 0.15, 0.20, or 0.25), we aim to remove $K = \lfloor \rho \times L \rfloor$ layers. Our selection strategy proceeds as follows:

\textbf{Triplet Selection.} We sort all triplets $\{\mathcal{T}_i\}$ in descending order by their similarity scores $S_{\text{triplet}}(i)$. Starting from the highest similarity triplet, we iteratively select triplets for processing. A triplet is only selected if none of its constituent layers have been previously assigned for dropping, fine-tuning, or freezing. This ensures each layer receives a single, well-defined treatment.

\textbf{Layer Assignment within Triplets.} For each selected triplet $\mathcal{T}_i = \{\ell_i, \ell_{i+1}, \ell_{i+2}\}$:

\begin{itemize}
\item \textit{Freeze}: Layer $\ell_{i+2}$ (the last layer in the triplet) is frozen to act as \textit{feature anchors} during fine-tuning.
\item \textit{Drop}: Among layers $\{\ell_i, \ell_{i+1}\}$, we drop the layer with higher $S_{\text{layer}}$ (more redundant).
\item \textit{fine-tune}: The remaining layer among $\{\ell_i, \ell_{i+1}\}$ is marked for fine-tuning.
\end{itemize}

The rationale for this assignment is threefold. First, freezing the last layer in each triplet constrains the fine-tuned layer to produce outputs compatible with downstream processing, preventing catastrophic distribution shift. Second, dropping the more redundant of the first two layers removes minimal information while preserving crucial transformations. Third, fine-tuning the less redundant layer enables it to learn compensatory representations, effectively performing the work of two layers. This process continues until we have selected $K$ layers for dropping, which also results in $K$ layers for fine-tuning and $K$ layers frozen. All remaining layers not involved in any selected triplet are frozen by default. Algorithm \ref{alg:interlace} formalizes this procedure.

\begin{algorithm}[t]
\caption{\textsc{Interlace} Layer Selection}
\label{alg:interlace}
\begin{algorithmic}[1]
\REQUIRE Model with $L$ layers, pruning ratio $\rho$, calibration dataset $\mathcal{D}$
\ENSURE Sets $\mathcal{L}_{\text{drop}}$, $\mathcal{L}_{\text{tune}}$, $\mathcal{L}_{\text{freeze}}$
\STATE Compute $S_{\text{layer}}(\ell)$ for all $\ell \in [1, L]$ using $\mathcal{D}$
\STATE Compute $S_{\text{triplet}}(i)$ for all $i \in [1, L-2]$ using $\mathcal{D}$
\STATE Sort triplets by $S_{\text{triplet}}$ in descending order: $\{\mathcal{T}_{\pi(1)}, \ldots, \mathcal{T}_{\pi(L-2)}\}$
\STATE Initialize $\mathcal{L}_{\text{drop}} = \emptyset$, $\mathcal{L}_{\text{tune}} = \emptyset$, $\mathcal{L}_{\text{freeze}} = \emptyset$
\STATE $K \gets \lfloor \rho \times L \rfloor$, $\text{assigned} \gets \emptyset$
\FOR{$k = 1$ to $L-2$}
    \STATE $\mathcal{T} \gets \{\ell_i, \ell_{i+1}, \ell_{i+2}\}$ where $i = \pi(k)$
    \IF{$\mathcal{T} \cap \text{assigned} = \emptyset$ AND $|\mathcal{L}_{\text{drop}}| < K$}
        \STATE $\mathcal{L}_{\text{freeze}} \gets \mathcal{L}_{\text{freeze}} \cup \{\ell_{i+2}\}$
        \IF{$S_{\text{layer}}(\ell_i) > S_{\text{layer}}(\ell_{i+1})$}
            \STATE $\mathcal{L}_{\text{drop}} \gets \mathcal{L}_{\text{drop}} \cup \{\ell_i\}$
            \STATE $\mathcal{L}_{\text{tune}} \gets \mathcal{L}_{\text{tune}} \cup \{\ell_{i+1}\}$
        \ELSE
            \STATE $\mathcal{L}_{\text{drop}} \gets \mathcal{L}_{\text{drop}} \cup \{\ell_{i+1}\}$
            \STATE $\mathcal{L}_{\text{tune}} \gets \mathcal{L}_{\text{tune}} \cup \{\ell_i\}$
        \ENDIF
        \STATE $\text{assigned} \gets \text{assigned} \cup \mathcal{T}$
    \ENDIF
\ENDFOR
\STATE $\mathcal{L}_{\text{freeze}} \gets \mathcal{L}_{\text{freeze}} \cup ([1,L] \setminus \text{assigned})$
\RETURN $\mathcal{L}_{\text{drop}}$, $\mathcal{L}_{\text{tune}}$, $\mathcal{L}_{\text{freeze}}$
\end{algorithmic}
\end{algorithm}

\subsection{Efficient Fine-Tuning Strategy}

After layer selection, we construct the pruned model by removing layers in $\mathcal{L}_{\text{drop}}$ and renumbering the remaining layers. The vision encoder, projection module, and embedding layers remain unchanged. During training, we update only the parameters of layers in $\mathcal{L}_{\text{tune}}$ while keeping all other components frozen. We employ standard cross-entropy loss for next-token prediction:

\begin{equation}
\mathcal{L} = -\sum_{t=1}^{T} \log p(y_t | y_{<t}, \mathbf{X})
\end{equation}

\begin{table*}[t]
\centering
\caption{Relative performance for \textsc{Interlace} across different layer dropping ratios with chain-of-thought reasoning enabled. All pruned models were fine-tuned for only \textbf{1 epoch} on just \textbf{1\% of the dataset}, updating only \textbf{a subset of layers} equivalent in number to those dropped. Values show (Dropped Score / Baseline Score × 100\%). For benchmarks with $\dagger$, we used scores reported in \cite{yang2025qwen3} as the baseline.}
\label{tab:main_results}
\setlength{\tabcolsep}{2pt}
\begin{tabular}{lcccccccc}
\toprule
\multirow{2}{*}{\textbf{Benchmark}} & \multicolumn{4}{c}{\textbf{Qwen3-VL-8B}} & \multicolumn{4}{c}{\textbf{Qwen3-VL-4B}} \\
\cmidrule(lr){2-5} \cmidrule(lr){6-9}
& \textbf{10\% Drop} & \textbf{15\% Drop} & \textbf{20\% Drop} & \textbf{25\% Drop} & \textbf{10\% Drop} & \textbf{15\% Drop} & \textbf{20\% Drop} & \textbf{25\% Drop} \\
\midrule
\multirow{5}{*}{\rotatebox[origin=c]{90}{\textbf{Text / Chart}}} 
\hspace{6pt}AI2D$_\text{Test}^\dagger$ & 93.4 & 92.5 & 88.5 & 84.4 & 94.3 & 90.6 & 85.0 & 79.7 \\
\hspace{16pt}ChartQA$_\text{Test}$ & 98.4 & 91.6 & 89.0 & 86.5 & 95.9 & 92.5 & 89.3 & 84.9 \\
\hspace{16pt}OCRBench$^\dagger$ & 93.2 & 89.6 & 87.1 & 86.8 & 90.9 & 89.1 & 85.6 & 80.0 \\
\hspace{16pt}TextVQA$_\text{Val}$ & 96.3 & 95.8 & 81.3 & 89.7 & 94.6 & 93.6 & 92.3 & 89.1 \\
\cmidrule(lr){2-9}
\hspace{16pt}\textbf{Average} & \textbf{\fpeval{round((93.4+98.4+93.2+96.3)/4, 1)}} & \textbf{\fpeval{round((92.5+91.6+89.6+95.8)/4, 1)}} & \textbf{\fpeval{round((88.5+89+87.1+81.3)/4, 1)}} & \textbf{\fpeval{round((84.4+86.5+86.8+89.7)/4, 1)}} & \textbf{\fpeval{round((94.3+95.9+90.9+94.6)/4, 1)}} & \textbf{\fpeval{round((90.6+92.5+89.1+93.6)/4, 1)}} & \textbf{88.0} & \textbf{\fpeval{round((79.7+84.9+80+89.1)/4, 1)}} \\
\multicolumn{7}{l}{ } \\
\multirow{4}{*}{\rotatebox[origin=c]{90}{\textbf{GVQA}}} 
\hspace{5pt}MMBench$_\text{DEV-EN-V1.1}^\dagger$ & 93.2 & 90.0 & 86.0 & 83.0 & 89.9 & 90.1 & 83.5 & 77.0 \\
\hspace{16pt}POPE & 99.5 & 99.6 & 98.9 & 99.5 & 99.3 & 99.1 & 98.7 & 98.2 \\
\hspace{16pt}RealWorldQA$^\dagger$ & 94.5 & 92.1 & 86.7 & 87.2 & 93.9 & 94.0 & 90.2 & 82.2 \\
\cmidrule(lr){2-9}
\hspace{16pt}\textbf{Average} & \textbf{\fpeval{round((93.2+99.5+94.5)/3, 1)}} & \textbf{\fpeval{round((90+99.6+92.1)/3, 1)}} & \textbf{\fpeval{round((86+98.9+86.7)/3, 1)}} & \textbf{\fpeval{round((83+99.5+87.2)/3, 1)}} & \textbf{\fpeval{round((89.9+99.3+93.9)/3, 1)}} & \textbf{\fpeval{round((90.1+99.1+94)/3, 1)}} & \textbf{\fpeval{round((83.5+98.7+90.2)/3, 1)}} & \textbf{\fpeval{round((77+98.2+82.2)/3, 1)}} \\
\multicolumn{7}{l}{ } \\
\multirow{4}{*}{\rotatebox[origin=c]{90}{\textbf{Perception}}} 
\hspace{5pt}HRBench4K$^\dagger$ & 91.4 & 92.1 & 87.1 & 84.0 & 93.6 & 91.0 & 88.8 & 80.1 \\
\hspace{16pt}HRBench8K$^\dagger$ & 87.1 & 90.0 & 84.1 & 82.1 & 92.8 & 92.1 & 87.3 & 81.8 \\
\hspace{16pt}VStar$^\dagger$ & 90.9 & 89.7 & 78.2 & 81.8 & 94.8 & 88.9 & 81.7 & 76.5 \\
\cmidrule(lr){2-9}
\hspace{16pt}\textbf{Average} & \textbf{\fpeval{round((91.4 + 87.1 + 90.9)/3, 1)}} & \textbf{\fpeval{round((92.1 + 90 + 89.7)/3, 1)}} & \textbf{\fpeval{round((87.1+84.1+78.2)/3, 1)}} & \textbf{\fpeval{round((84+82.1+81.8)/3, 1)}} & \textbf{\fpeval{round((93.6+92.8+94.8)/3, 1)}} & \textbf{\fpeval{round((91+92.1+88.9)/3, 1)}} & \textbf{\fpeval{round((88.8+87.3+81.7)/3, 1)}} & \textbf{\fpeval{round((80.1+81.8+76.5)/3, 1)}} \\
\multicolumn{7}{l}{ } \\
\multirow{3}{*}{\rotatebox[origin=c]{90}{\textbf{Inst\&Sci}}} 
\hspace{7pt}MIABench$^\dagger$ & 95.1 & 89.6 & 88.9 & 85.0 & 92.7 & 89.7 & 87.0 & 77.9 \\
\hspace{16pt}ScienceQA$_\text{Test}$ & 94.6 & 92.9 & 86.6 & 83.5 & 94.4 & 92.6 & 86.2 & 72.6 \\
\cmidrule(lr){2-9}
\hspace{16pt}\textbf{Average} & \textbf{\fpeval{round((95.1+94.6)/2, 1)}} & \textbf{\fpeval{round((89.6+92.9)/2, 1)}} & \textbf{\fpeval{round((88.9+86.6)/2, 1)}} & \textbf{\fpeval{round((85+83.5)/2, 1)}} & \textbf{\fpeval{round((94.4+92.7)/2, 1)}} & \textbf{\fpeval{round((92.6+89.7)/2, 1)}} & \textbf{\fpeval{round((86.2+87)/2, 1)}} & \textbf{\fpeval{round((72.6+77.9)/2, 1)}} \\
\midrule
\midrule
\hspace{12pt}\textbf{Overall} & \textbf{94.0} & \textbf{\fpeval{round((92.5+91.6+89.6+95.8+90+99.6+92.1+92.1+90+89.7+89.6+92.9)/12, 1)}} & \textbf{\fpeval{round((88.5+89+87.1+81.3+86+98.9+86.7+87.1+84.1+78.2+88.9+86.6)/12, 1)}} & \textbf{\fpeval{round((84.4+86.5+86.8+89.7+83+99.5+87.2+84+82.1+81.8+85+83.5)/12, 1)}} & \textbf{\fpeval{round((94.3+95.9+90.9+94.6+89.9+99.3+93.9+93.6+92.8+94.8+92.7+94.4)/12, 1)}} & \textbf{\fpeval{round((90.6+92.5+89.1+93.6+90.1+99.1+94+91+92.1+88.9+89.7+92.6)/12, 1)}} & \textbf{\fpeval{round((85+89.3+85.6+92.3+83.5+98.7+90.2+88.8+87.3+81.7+87+86.2)/12, 1)}} & \textbf{\fpeval{round((79.7+84.9+80+89.1+77+98.2+82.2+80.1+81.8+76.5+77.9+72.6)/12, 1)}} \\[4pt]
\bottomrule
\end{tabular}
\end{table*}

where $T$ is the sequence length and $y_t$ are the target tokens. This simple objective proves sufficient for recovering performance, as our strategic layer selection supports fast convergence. As for training configurations, we use a learning rate of $1 \times 10^{-5}$ with cosine decay, batch size of 16 with 2 gradient accumulation steps, and train for a single epoch on 1\% of the FineVision dataset \cite{wiedmann2025finevision}. FineVision comprises over 200 diverse datasets with 24 million samples spanning general VQA, document understanding, OCR, science reasoning, and agentic tasks. We select 1\% uniformly from each constituent dataset, yielding approximately 240K training samples. To further reduce memory consumption, we filter samples with more than 3 images or exceeding 5 turns of conversation. This aggressive data and training regime demonstrates the sample efficiency of our approach, as the frozen layers provide strong inductive biases that accelerate convergence.

\section{Experiments}
\subsection{Experimental Setup}
We evaluate \textsc{Interlace} on Qwen3-VL-Instruct models with 8B and 4B parameters~\cite{yang2025qwen3}. Both architectures consist of 36 transformer layers in their language components, with hidden state dimensions of 4096 and 2560 for the 8B and 4B versions respectively. Our evaluation spans 12 diverse benchmarks organized into four task categories to comprehensively assess multimodal reasoning, instruction following, spatial understanding, scientific knowledge, optical character recognition, and cross-modal alignment capabilities. For text and chart understanding visual tasks, we include AI2D~\cite{kembhavi2016diagram}, ChartQA~\cite{masry2022chartqa}, OCRBench~\cite{liu2024ocrbench}, and TextVQA~\cite{singh2019towards}. For general visual question answering, we evaluate on MMBench$_{\text{DEV-EN-V1.1}}$~\cite{liu2024mmbench}, RealWorldQA~\cite{zhang2024mme}, and POPE~\cite{li2023evaluating}. We assess fine-grained visual perception with V$^{*}$~\cite{cheng2025v}, HRBench4K and HRBench8K~\cite{wang2025divide} benchmarks. Finally, science-related reasoning and instruction following capabilities are measured using ScienceQA~\cite{lu2022learn} and MIABench~\cite{qian2024mia}.

We use VLMEvalKit~\cite{duan2024vlmevalkit} for all benchmark evaluations, following standard evaluation protocols for each task. For some baseline model scores in Table~\ref{tab:main_results}, we rely on results reported in the Qwen3 Technical Report~\cite{yang2025qwen3}. However, for benchmarks not included in their report, specifically POPE, ScienceQA, ChartQA, and TextVQA, we evaluate the baseline models ourselves using VLMEvalKit with the recommended procedures from the respective benchmark authors. We report relative performance as the ratio of pruned model score to baseline model score, expressed as a percentage.

All experiments are conducted on NVIDIA RTX PRO 6000 Blackwell Workstation Edition GPUs with 98GB of memory. The post-pruning fine-tuning phase takes approximately six hours per model on 1\% of the dataset with 240,000 samples for one epoch. Memory consumption scales with the pruning ratio and model size, as higher pruning ratios necessitate fine-tuning more layers, and the 8B model's larger hidden state dimension increases memory requirements for gradient computation and optimizer states. To manage these demands, we employ mixed-precision training using bfloat16 arithmetic and gradient checkpointing, which trades computation for memory by recomputing activations during the backward pass.

\subsection{Main Results}

Table~\ref{tab:main_results} presents the performance of \textsc{Interlace} across four different layer dropping ratios ranging from 10\% to 25\% on both the 8B and 4B parameter models. To establish meaningful performance baselines, we compare pruned models against the original unpruned versions to quantify the performance-efficiency trade-off at different compression rates. For these evaluations, we allow both baseline and pruned models to perform Chain-Of-Thought (COT) reasoning when beneficial by setting the maximum number of generation tokens to 16384, ensuring a fair comparison that does not artificially handicap either model's reasoning capabilities. The results demonstrate that \textsc{Interlace} achieves remarkable performance retention across all benchmark categories. For the 8B model, dropping 10\% of layers maintains 94.0\% of baseline performance on average, while even at 25\% layer removal, we retain 86.1\% of the original performance. The performance degradation is notably smooth across pruning ratios, with the model maintaining strong capabilities in text and chart understanding tasks even at higher compression rates. The 4B model exhibits similar trends, achieving 93.9\% average performance retention at 10\% pruning and 81.7\% at 25\% pruning. 

The breakdown across task categories reveals interesting patterns in how different capabilities degrade under compression. General visual question answering tasks show the most robust performance retention, with the 8B model maintaining 89.9\% of baseline performance even at 25\% pruning. Text and chart understanding tasks follow closely, retaining 86.8\% at the same compression rate. Fine-grained perception tasks, which require detailed spatial reasoning, show more sensitivity to layer removal but still maintain 82.6\% performance at 25\% pruning. This degradation pattern aligns with our intuition that higher-level reasoning capabilities can be preserved through strategic layer selection and targeted fine-tuning, while low-level visual perception may require more extensive architectural support.


\begin{center}
    \centering
    \includegraphics[width=0.47\textwidth]{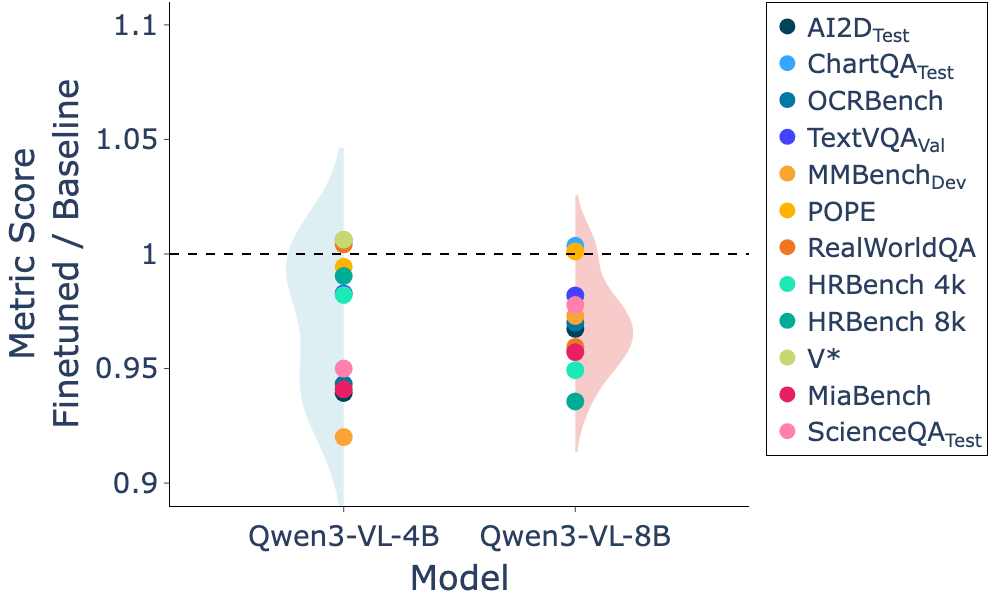}
    \captionof{figure}{\textbf{Effect of fine-tuning on baseline performance with COT enabled.} Benchmarks below the dashed line at 1.0 show performance degradation after fine-tuning without layer pruning. }
    \label{fig:FT_Effect}
\end{center}

\subsection{Isolating Fine-Tuning Contributions}

\begin{table*}[t]
\centering
\caption{Comparing different pruning methods across benchmark groups in Qwen3-VL-8B. RP is the average relative performance compared to fine-tuned dense baseline model (Dense-FT). Reported results are after limiting chain-of-thought reasoning to 50 tokens.}
\label{tab:compare}
\begin{tabular}{l c c c | cccc | cc}
\toprule
\multirow{2}{*}{\textbf{Method}} & 
\multirow{2}{*}{\textbf{Sparsity}} & 
\multirow{2}{*}{\textbf{Fine-Tune}} & 
\multirow{2}{*}{\textbf{TTFT}} & 
\multicolumn{4}{c|}{$\uparrow$ \textbf{Performance on Benchmark Group (\%)}} & 
\multirow{2}{*}{\textbf{Avg}} &
\multirow{2}{*}{\textbf{RP}} \\
\cmidrule(lr){5-8}
 & & & & \textbf{Text/Chart} & \textbf{GVQA} & \textbf{Perception} & \textbf{Inst\&Sci} \\
\midrule
Dense & 0\% & \ding{55} & 1.00$\times$ & 79.3 & 79.1 & 76.5 & 74.9 & 77.8 & 97.1 \\
Dense-FT & 0\% & \ding{51} & 1.00$\times$ & 83.2 & 80.2 & 75.8 & 82.4 & 80.5 & 100.0 \\
\midrule
\midrule
Wanda 2:4 & 50\% & \ding{55} & 0.97$\times$ & 6.1 & 7.8 & 5.7 & 10.7 & 7.2& 8.9 \\
Magnitude 2:4 & 50\% & \ding{55} & 0.97$\times$ & 6.2 & 7.6 & 7.9 & 10.6 & 7.7 & 9.5 \\
\midrule
SLEB & 25\% & \ding{55} & 1.12$\times$ & 43.4 & 54.1 & 48.4 & 51.3 & 48.6 & 60.5 \\
SLEB-FT & 25\% & \ding{51} & 1.12$\times$ & 50.5 & 43.8 & 41.4 & 47.4 & 46.0 & 57.1 \\
\midrule
Interlace (Ours) & 25\% & \ding{51} & 1.18$\times$ & \textbf{74.5} & \textbf{73.6} & \textbf{64.9} & \textbf{72.8} & \textbf{71.6} & \textbf{88.9} \\
\bottomrule
\end{tabular}
\end{table*}

To rigorously isolate the contribution of our triplet-based selection strategy from the benefits of the training dataset itself, we conduct a controlled experiment examining the effect of fine-tuning on the unpruned baseline model. Specifically, we fine-tune the last 25\% of layers in the 8B model on the same 1\% FineVision subset for the same single epoch, matching the exact training configuration used for our pruned models. This ablation allows us to determine whether performance improvements stem from our strategic layer selection or simply from additional training on high-quality data. Figure~\ref{fig:FT_Effect} presents the results of this analysis through violin plots showing the distribution of relative performance changes across all benchmarks. The horizontal line at $y=1.0$ represents baseline performance, with points below indicating degradation and points above indicating improvement. The results reveal that fine-tuning without pruning actually degrades performance on most benchmarks compared to the baseline, with the majority of tasks falling below the baseline threshold. This counterintuitive result demonstrates that naively fine-tuning a subset of layers on limited data, even high-quality data, can disrupt the carefully balanced representations learned during pre-training and direct policy optimization phases. This controlled experiment validates a key hypothesis underlying \textsc{Interlace}: our method's success does not merely stem from exposure to additional training data, but rather from the synergistic combination of strategic layer removal, targeted fine-tuning, and architectural constraints through frozen anchor layers. The structural changes induced by pruning create a training landscape where fine-tuning on minimal data becomes beneficial rather than harmful. These results were obtained with the same chain-of-thought reasoning setup used in our main results, with the maximum generation tokens set to 16384.

\subsection{Comparison with Alternative Pruning Methods}

Table~\ref{tab:compare} presents a comprehensive comparison of \textsc{Interlace} against several competitive baselines on the 8B model. We evaluate against the original unpruned dense model, Wanda~\cite{sun2023simple} and magnitude-based pruning~\cite{back2025magnitude} with 2:4 pruning patterns, and SLEB~\cite{song2024sleb} with 25\% layer dropping rate. For this comparison, we limit all models to generate a maximum of 50 tokens per input, as some heavily pruned models struggle to predict end tokens due to their reduced capacity. Additionally, to ensure a fair comparison, we include results from fine-tuned versions of both the dense baseline and SLEB, training the same number of layers on identical data for the same duration as \textsc{Interlace}. We denote these variants as Dense-FT and SLEB-FT respectively.

The results demonstrate that \textsc{Interlace} substantially outperforms all pruning methods across all benchmark categories. Comparing against the original dense model, which achieves 77.8\% average performance, our method with 25\% pruning reaches 71.6\% while providing a 1.18× speedup in Time-To-First-Token (TTFT). More critically, when comparing against Dense-FT, which represents the upper bound of what fine-tuning alone can achieve without compression (80.5\% average), \textsc{Interlace} demonstrates that we can compress the model significantly while retaining 88.9\% of the fine-tuned dense model's performance with inference acceleration.

The comparison with weight pruning methods reveals the fundamental limitations of 2:4 approaches for vision-language models. Both Wanda and magnitude-based pruning with 2:4 patterns achieve only 7.2\% and 7.7\% average performance respectively, despite requiring specialized hardware for actual speedup realization. These performance drops suggest that the coupled nature of vision-language representations makes them particularly vulnerable to weight-level sparsity, where structured patterns like 2:4 fail to preserve critical cross-modal dependencies.

SLEB, which also performs layer-level pruning, achieves 48.6\% average performance without fine-tuning, demonstrating that layer removal alone causes substantial capability loss. Interestingly, SLEB-FT performs worse than the unfine-tuned SLEB variant at 46.0\%, echoing our earlier finding that naive fine-tuning strategies can degrade performance. This suggests that SLEB's consecutive layer dropping strategy creates training instabilities that prevent effective recovery through standard fine-tuning. In contrast, \textsc{Interlace} \textbf{achieves 88.9\%} average relative performance, showcasing a \textbf{28.4\% improvement} over SLEB and a \textbf{31.8\% improvement} over SLEB-FT. 


\subsection{Ablation Studies}

To rigorously validate each component of our triplet-based layer selection methodology, we implement several alternative layer selection and training strategies, applying identical training procedures to each approach to ensure a fair comparison. Table~\ref{tab:ablations} presents the results of these ablation studies, with all experiments performed on the Qwen3-VL-8B model at 25\% layer dropping rate with chain-of-thought reasoning enabled. For each ablation, we fine-tune an equivalent number of layers on the same dataset for the same number of epochs, isolating the impact of the selection strategy itself.

\textbf{Consecutive Dropping and Fine-Tuning.} This baseline implements the most straightforward pruning strategy: we calculate the average cosine similarity of tokens for individual layers on 10\% of the fine-tuning dataset (24,000 samples), identify the most redundant consecutive block of layers based on cumulative similarity scores, drop this entire block, and fine-tune an equivalent number of consecutive layers placed immediately after the dropped block to compensate for lost capacity. This approach mimics previous structured pruning methods that remove contiguous architectural components. The results demonstrate the lowest performance, achieving an average relative performance of 65.1\% relative to \textsc{Interlace}. This validates our core hypothesis that large blocks of modified consecutive layers create training instabilities in vision-language models, preventing effective fine-tuning even with our carefully curated dataset and extended reasoning capabilities.

\textbf{Random Layer Selection.} To establish whether our similarity-based selection provides genuine benefits over random pruning, we randomly select 25\% of layers for dropping and randomly select an equal number from the remaining layers for fine-tuning, keeping all other layers frozen. This ablation controls for the benefits of our interleaved freeze-finetune pattern while removing the intelligence of our selection process. Interestingly, random selection outperforms consecutive dropping by a margin of 20\% (85.1\% versus 65.1\% average), further confirming that avoiding large consecutive blocks of modifications is crucial for fast divergence, and that even random interleaving provides substantial benefits over structured consecutive removal.

\begin{table}[t]
\centering
\caption{Ablation studies showing relative performance (compared to Interlace = 100\%) for different layer selection and training strategies on Qwen3-VL-8B with 25\% pruning and COT enabled.}
\label{tab:ablations}
\setlength{\tabcolsep}{2pt}
\begin{tabular}{lcccc | c}
\toprule
\textbf{Method} & \textbf{Text/Chart} & \textbf{GVQA} & \textbf{Perc.} & \textbf{Inst\&Sci} & \textbf{Avg} \\
\midrule
Consecutive & 76.6 & 55.9 & 54.8 & 71.4 & 65.1 \\
Random & 87.9 & 86.9 & 77.5 & 87.9 & 85.1 \\
Interlace-OA & 95.8 & 96.2 & 89.6 & 98.9 & 94.9 \\
Interlace-TN & 99.4 & 97.3 & 99.3 & 98.3 & 98.7 \\
\bottomrule
\end{tabular}
\end{table}

\textbf{Interlace-Ordered-Assignment (Interlace-OA).} This variant tests whether individual layer analysis within triplets provides value beyond triplet-level redundancy detection. We perform the same triplet redundancy analysis and select the most redundant triplets, but skip the individual layer similarity calculations within each triplet. Instead, we assign layer roles purely by position: the first layer is always dropped, the second is always fine-tuned, and the third is always frozen. This tests whether our dual-level analysis (both triplet and individual layer metrics) is necessary or if the triplet structure alone suffices. Interlace-OA achieves an average relative performance of 94.9\% across all tasks, validating the importance of our triplet-based approach. However, it still falls short in comparison to \textsc{Interlance}, indicating that the individual layer analysis within triplets provides consistent improvements across all benchmarks.

\textbf{Interlace-Train-Next (Interlace-TN).} This ablation tests the importance of our triplet analysis by using only individual layer similarity scores. We start from the most redundant layer and iteratively drop layers in order of redundancy, fine-tuning the layer located immediately after each dropped layer, and freezing all remaining layers once 25\% of layers are dropped. This approach maintains an interleaved pattern of dropped and fine-tuned layers, but critically, it does not guarantee that frozen layers are strategically placed after fine-tuning layers to act as stable anchors. Whether frozen layers emerge between fine-tuning layers depends entirely on the distribution of similarity scores across the network. If high-similarity layers are clustered together, multiple consecutive fine-tuning layers may occur without intervening frozen anchors. Conversely, if redundancy is spatially distributed, frozen layers may naturally intersperse with fine-tuned ones. This architectural uncertainty makes the method's effectiveness highly dependent on the specific pretrained network's redundancy structure, potentially limiting its generalizability across different model families or training regimes. Despite this lack of architectural control, Interlace-TN achieves a 98.7\% average performance, standing below our main \textsc{Interlace} model. These results suggest that for this particular architecture, redundant layers are sufficiently distributed that frozen anchors mostly emerge in beneficial positions. However, the full \textsc{Interlace} method's explicit triplet-based placement of frozen anchors provides guaranteed architectural stability and more predictable performance across diverse model architectures, making it the more reliable choice for general deployment scenarios where robust performance must be ensured regardless of the underlying redundancy patterns.

\section{Conclusion}

We introduced \textsc{Interlace}, a novel framework for efficient compression of large vision-language models through strategic layer pruning and sample-efficient fine-tuning. By analyzing triplets of consecutive layers to identify local redundancy patterns and implementing an interleaved freeze-finetune architecture, our method achieves remarkable performance retention while requiring only minimal training data. Our comprehensive evaluation demonstrates that with chain-of-thought reasoning enabled, \textsc{Interlace} retains 94.0\% of baseline performance after removing 10\% of layers and 86.1\% after removing 25\% of layers, substantially outperforming alternative pruning approaches while delivering off-the-shelf inference acceleration. The success of our triplet-based selection strategy and constrained fine-tuning paradigm suggests that structured architectural modifications, when combined with strategic training constraints, offer a promising path toward deploying high-capability vision-language models in resource-constrained environments. Future work may explore extending this framework to other multimodal architectures, investigating adaptive pruning ratios tailored to specific deployment scenarios, and studying the effect of the size of the dataset on different pruning methods and tasks.

{
    \small
    \bibliographystyle{ieeenat_fullname}
    \bibliography{main}
}

\clearpage
\setcounter{page}{1}
\maketitlesupplementary

\appendix
\label{sec:appendix}



\section{Appendix}

\subsection{Per-Benchmark Inference Speedup Analysis}

To provide a comprehensive understanding of the computational benefits achieved by our method, we report detailed Time-To-First-Token (TTFT) speedup measurements across all twelve evaluation benchmarks in Table~\ref{tab:ttft_speedup}. These measurements were obtained by averaging TTFT across all samples in each benchmark using Qwen3-VL 8B and 4B models with 10 to 25\% layer pruning ratios. As the pruning ratio increases to 15\% and beyond, all benchmarks exhibit clear acceleration benefits. Fine-grained high-resolution benchmarks such as HRBench4K and HRBench8K show more modest speedup gains across pruning ratios compared to other tasks, which we attribute to the computational bottleneck shifting toward vision encoding and projection operations rather than language model inference in these vision-intensive scenarios. Conversely, text-heavy benchmarks like AI2D, ChartQA, and ScienceQA demonstrate the highest speedup ratios at 25\% pruning, reaching up to 1.22$\times$ acceleration, as their longer token sequences amplify the benefits of reduced layer computations.

\begin{table*}[t]
\centering
\small
\caption{Time-To-First-Token (TTFT) speedup for INTERLACE across different pruning ratios and model sizes. All speedup factors are relative to the unpruned baseline models.}
\label{tab:ttft_speedup}
\setlength{\tabcolsep}{13pt}
\begin{tabular}{llcccccccc}
\toprule
& & \multicolumn{4}{c}{Qwen3-VL-8B} & \multicolumn{4}{c}{Qwen3-VL-4B} \\
\cmidrule(lr){3-6} \cmidrule(lr){7-10}
Category & Benchmark & 10\% & 15\% & 20\% & 25\% & 10\% & 15\% & 20\% & 25\% \\
\midrule
\multirow{5}{*}{Text/Chart} 
& AI2D & 0.98 & 1.11 & 1.15 & 1.21 & 0.96 & 1.10 & 1.14 & 1.19 \\
& ChartQA & 0.97 & 1.11 & 1.16 & 1.21 & 0.95 & 1.09 & 1.14 & 1.19 \\
& OCRBench & 1.02 & 1.08 & 1.11 & 1.15 & 1.00 & 1.08 & 1.11 & 1.14 \\
& TextVQA & 1.02 & 1.10 & 1.15 & 1.20 & 0.99 & 1.09 & 1.14 & 1.19 \\
\cmidrule(lr){2-10}
& Average & 1.00 & 1.10 & 1.14 & 1.19 & 0.98 & 1.09 & 1.13 & 1.18 \\
\midrule
\multirow{4}{*}{GVQA}
& MMBench & 0.96 & 1.11 & 1.16 & 1.22 & 0.95 & 1.10 & 1.14 & 1.19 \\
& POPE & 0.95 & 1.11 & 1.16 & 1.22 & 0.95 & 1.10 & 1.14 & 1.20 \\
& RealWorldQA & 1.01 & 1.09 & 1.14 & 1.18 & 0.98 & 1.09 & 1.13 & 1.17 \\
\cmidrule(lr){2-10}
& Average & 0.97 & 1.10 & 1.15 & 1.21 & 0.96 & 1.10 & 1.14 & 1.19 \\
\midrule
\multirow{4}{*}{Perception}
& HRBench4K & 1.03 & 1.05 & 1.08 & 1.10 & 1.02 & 1.06 & 1.08 & 1.10 \\
& HRBench8K & 1.03 & 1.05 & 1.08 & 1.10 & 1.03 & 1.05 & 1.08 & 1.10 \\
& VStar & 1.06 & 1.10 & 1.13 & 1.17 & 1.00 & 1.08 & 1.11 & 1.14 \\
\cmidrule(lr){2-10}
& Average & 1.04 & 1.07 & 1.10 & 1.12 & 1.02 & 1.06 & 1.09 & 1.11 \\
\midrule
\multirow{3}{*}{Inst\&Sci}
& MIABench & 1.02 & 1.07 & 1.10 & 1.13 & 1.00 & 1.07 & 1.10 & 1.13 \\
& ScienceQA & 0.96 & 1.12 & 1.16 & 1.22 & 0.95 & 1.10 & 1.15 & 1.20 \\
\cmidrule(lr){2-10}
& Average & 0.99 & 1.09 & 1.13 & 1.18 & 0.98 & 1.08 & 1.12 & 1.17 \\
\midrule
\multicolumn{2}{l}{\textbf{Overall}} 
& \textbf{1.00} & \textbf{1.09} & \textbf{1.14} & \textbf{1.18} 
& \textbf{0.99} & \textbf{1.08} & \textbf{1.12} & \textbf{1.17} \\
\bottomrule
\end{tabular}
\end{table*}

\subsection{Comprehensive Ablation Study Results}

Table~\ref{tab:ablation_full} presents the complete performance breakdown of our ablation studies across all twelve benchmarks, extending the aggregated results shown in Table \ref{tab:ablations}. Each ablation variant was evaluated under identical conditions with the 8B model at 25\% pruning ratio and chain-of-thought reasoning enabled. The detailed absolute performance scores reveal several critical patterns in how different layer selection strategies affect specific task categories. The consecutive dropping baseline exhibits catastrophic degradation on general visual question answering tasks, achieving only 8.3\% on MMBench compared to the 70.5\% achieved by INTERLACE, which represents a striking 62.2 percentage point gap. This severe collapse suggests that vision-language alignment and cross-modal reasoning are especially vulnerable to large blocks of consecutive architectural modifications. Perception-intensive tasks such as HRBench4K, HRBench8K, and VStar also suffer dramatically under consecutive pruning, losing over 30 absolute percentage points compared to INTERLACE. Random layer selection substantially improves upon consecutive dropping, with MMBench performance jumping to 53.0\%, but still exhibits significant gaps across all benchmarks. The Interlace-OA variant, which uses fixed positional assignment within triplets, achieves competitive performance across most tasks, reaching 65.9\% on AI2D and 85.9\% on POPE, demonstrating that the triplet structure alone provides substantial benefits. Interlace-TN performs remarkably close to the full INTERLACE method across nearly all benchmarks, with particularly strong results on perception tasks where it achieves 68.5\% on HRBench4K and 64.9\% on VStar. This validates the effectiveness of individual layer importance scoring while highlighting that our complete triplet-based framework with explicit anchor placement provides the most consistent and predictable performance across the diverse benchmark suite.


\begin{table*}[t]
\centering
\small
\caption{Complete ablation study results for Qwen3-VL 8B with 25\% pruning rate. Results show absolute percentage performance values for each benchmark task with chain-of-thought reasoning enabled. Results are after one epoch of training on 1\% of FineVision dataset.}
\label{tab:ablation_full}
\setlength{\tabcolsep}{15pt}
\begin{tabular}{lccccc | c}
\toprule
Benchmark & Consecutive & Random & Interlace-OA & Interlace-TN & \textsc{Interlace} & Dense \\
\midrule
AI2D & 33.8 & 52.8 & 65.9 & 69.8 & \textbf{72.3} & 85.7	\\
ChartQA & 62.2 & 64.2 & 67.7 & \textbf{73.0} & 72.0 & 83.3 \\
OCRBench & 67.1 & 75.0 & 76.7 & 77.1 & \textbf{77.8} & 89.6 \\
TextVQA & 64.9 & 69.4 & 74.1 & \textbf{75.1} & 74.5 & 82.9 \\
MMBench & 8.3 & 53.0 & 65.6 & 67.6 & \textbf{70.5} & 85.0 \\
POPE & 75.8 & 85.8 & 85.9 & 86.5 & \textbf{87.7} & 88.0 \\
RealWorldQA & 43.3 & 54.8 & 60.9 & 60.8 & \textbf{62.4} & 71.5 \\
HRBench4K & 35.3 & 53.6 & 61.9 & \textbf{68.5} & 66.3 & 78.9 \\
HRBench8K & 35.4 & 49.4 & 54.4 & \textbf{63.0} & 61.3 & 74.6 \\
VStar & 37.7 & 50.3 & 61.3 & 64.9 & \textbf{70.7} & 86.4 \\
MIABench & 62.5 & 73.7 & \textbf{78.6} & 76.8 & 77.6 & 91.1	\\
ScienceQA & 49.0 & 63.4 & 76.0 & 76.8 & \textbf{78.7} & 94.2 \\
\midrule
Average & 47.9 & 62.1 & 69.1 & 71.7 & \textbf{72.7} & 84.3 \\
\bottomrule
\end{tabular}
\end{table*}

\subsection{Layer and Triplet Similarity Distribution Analysis}

\begin{figure*}[t]
\centering
\includegraphics[width=1\textwidth]{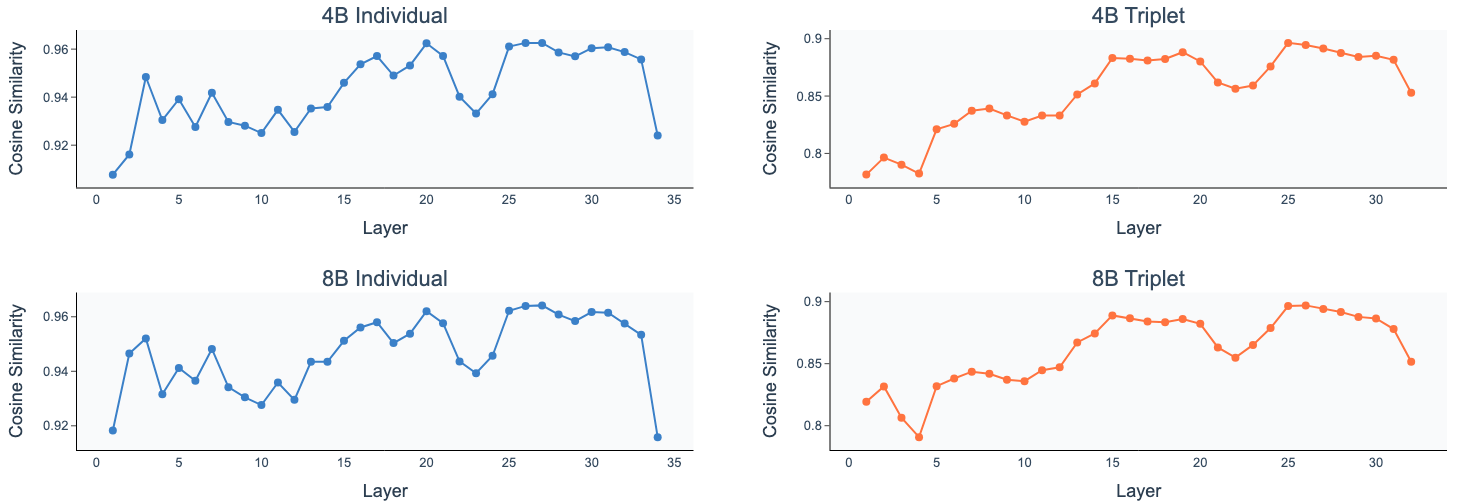}
\caption{Distribution of cosine similarity scores for individual layers and triplets across the depth of Qwen3-VL-8B (blue) and 4B (orange) models. Higher scores indicate greater redundancy.}
\label{fig:similarity_distribution}
\end{figure*}

Figure~\ref{fig:similarity_distribution} visualizes the distribution of cosine similarity scores for both individual layers and triplets across the entire depth of the Qwen3-VL-8B and 4B models. The similarity scores were computed using 10\% of the fine-tuning dataset (24,000 samples), measuring the average cosine similarity between hidden states before and after each layer or triplet. Both models exhibit higher redundancy in middle layers, with similarity scores peaking around layers 15-25, suggesting that these regions contain more functionally overlapping transformations that can be safely removed. The deeper layers beyond layer 30 show lower similarity scores, indicating more specialized and less redundant functionality that is critical for final output generation. The triplet similarity scores generally follow similar trends to individual layer scores but with smoother transitions.

\begin{table}[t]
\centering
\small
\caption{Effect of fine-tuning 25\% of layers in the unpruned dense baseline under enabled or limited Chain-of-Thought (CoT) reasoning. Dense-FT is fine-tuned on 1\% of the FineVision dataset for one epoch. Values denote absolute performance scores (\%).}
\label{tab:baseline_finetune}
\setlength{\tabcolsep}{6pt}
\begin{tabular}{lcccc}
\toprule
\multirow{2}{*}{Benchmark} & \multicolumn{2}{c}{CoT \ding{51}} & \multicolumn{2}{c}{CoT \ding{55}} \\
\cmidrule(lr){2-3} \cmidrule(lr){4-5}
& Dense & Dense-FT & Dense & Dense-FT \\
\midrule
AI2D        & 85.7 & 82.9 & 69.9 & 83.0 \\
ChartQA     & 83.3 & 83.6 & 83.3 & 83.4 \\
OCRBench    & 89.6 & 85.8 & 81.2 & 85.3 \\
TextVQA     & 82.9 & 81.4 & 82.9 & 81.0 \\
MMBench     & 85.0 & 82.7 & 77.9 & 82.1 \\
POPE        & 88.0 & 88.1 & 88.0 & 88.3 \\
RealWorldQA & 71.5 & 68.6 & 71.4 & 70.1 \\
HRBench4K   & 78.9 & 74.9 & 77.0 & 74.5 \\
HRBench8K   & 74.6 & 69.8 & 70.8 & 70.8 \\
VStar       & 86.4 & 82.7 & 81.7 & 82.2 \\
MIABench    & 91.1 & 87.2 & 85.0 & 81.7 \\
ScienceQA   & 94.2 & 92.1 & 75.2 & 91.7 \\
\midrule
Average     & 84.3 & 81.6 & 78.7 & 81.2 \\
\bottomrule
\end{tabular}
\end{table}

\subsection{Fine-Tuning Effects on Unpruned Baseline Models}

Table~\ref{tab:baseline_finetune} provides quantitative analysis of how fine-tuning affects the unpruned baseline model both with and without chain-of-thought reasoning, expanding upon the visualization presented in Figure \ref{fig:FT_Effect} of the main paper. We fine-tuned the last 25\% of layers in the Qwen3-VL-8B model on 1\% of FineVision for one epoch, matching the training configuration used for our pruned models. With chain-of-thought reasoning enabled by setting the maximum token generation to 16384, the unpruned dense baseline achieves an average of 84.3\% performance, while fine-tuning degrades this to 81.6\%, representing a 2.7 percentage point drop. When chain-of-thought reasoning is limited to 50 tokens, the pattern becomes more complex. The baseline performance drops to 78.7\% on average without extended reasoning capability, but fine-tuning actually improves performance to 81.2\%, suggesting that the fine-tuning process helps the model adapt to generate more concise responses. Notably, AI2D shows a dramatic improvement from 69.9\% to 83.0\% under limited CoT conditions after fine-tuning, while ScienceQA jumps from 75.2\% to 91.7\%. However, this improvement comes at the cost of reduced performance on several other benchmarks.

\subsection{Fine-Tuning Hyperparameter Configuration}

In our fine-tuning procedure, we use AdamW optimizer with a learning rate of 1$\times$10$^{-5}$. The learning rate follows a cosine annealing schedule with a warmup ratio of 0.03 (3\% of total training steps), which allows the optimizer to gradually adapt to the modified architecture. We explicitly set weight decay to 0 to avoid constraining the model's ability to rapidly adapt its representations during the short one-epoch training regime. We employ a per-device batch size of 16 with 2 gradient accumulation steps, yielding an effective batch size of 32, which fits within the memory constraints of our NVIDIA RTX PRO 6000 GPUs while maintaining stable gradient estimates. Mixed-precision training using bfloat16 arithmetic reduces memory consumption compared to full float32 training while preserving numerical stability. We use gradient clipping with a maximum norm of 1.0 to prevent occasional large gradients from destabilizing training, which is particularly important given that we are fine-tuning only a sparse subset of layers. We apply gradient checkpointing to all trainable layers. We utilize 4 dataloader workers and employ DeepSpeed ZeRO Stage 3 optimization with overlap communication and contiguous gradients enabled to further reduce memory footprint and enable training on consumer-grade hardware. 

\end{document}